\documentclass[preprint,12pt]{elsarticle}

\usepackage{amssymb}
\usepackage{amsmath}

\usepackage{xcolor}
\usepackage{booktabs}
\usepackage{multirow}
\usepackage{array}
\usepackage{hyperref}
\usepackage{makecell}
\usepackage{multicol}
\usepackage{adjustbox}
\usepackage{longtable}
\usepackage{caption}
\usepackage{subcaption}

\journal{Accident Analysis and Prevention}

\newcommand{\comment}[1]{ }

\graphicspath{{images/}}

\begin{document}

\begin{frontmatter}



\title{Predicting and Explaining Traffic Crash Severity Through Crash Feature Selection} 


\author[hri]{Andrea Castellani} 
\author[jad]{Zacharias Papadovasilakis}
\author[jad]{Giorgos Papoutsoglou}
\author[osu]{Mary Cole}
\author[hdma]{Brian Bautsch}
\author[hri]{Tobias Rodemann}
\author[jad,crete]{Ioannis Tsamardinos}
\author[osu]{Angela Harden}

\affiliation[hri]{organization={Honda Research Institute Europe},
            addressline={Carl-Legien-Straße 30}, 
            city={Offenbach},
            postcode={63073}, 
            state={Hesse},
            country={Germany}}
\affiliation[osu]{organization={The Ohio State University},
addressline={281 W Lane Ave}, 
city={Columbus},
postcode={43210}, 
state={Ohio},
country={United States}}
\affiliation[jad]{organization={JADBio Gnosis DA S.A.},
addressline={N. Plastira 100, Vasilika Vouton}, 
city={Heraklion},
postcode={700 13}, 
state={Crete},
country={Greece}}
\affiliation[crete]{organization={Department of Computer Science, University of Crete},
addressline={GR-70013}, 
city={Heraklion},
postcode={700 13}, 
state={Creete},
country={Greece}}
\affiliation[hdma]{organization={American Honda Motor Co., Inc.},
addressline={1919 Torrance Boulevard}, 
city={Torrance},
postcode={90501-2746}, 
state={California},
country={United States}}

\begin{abstract}
Motor vehicle crashes remain a leading cause of injury and death worldwide, necessitating data-driven approaches to understand and mitigate crash severity. This study introduces a curated dataset of more than 3 million people involved in accidents in Ohio over six years (2017-2022), aggregated to more than 2.3 million vehicle-level records for predictive analysis. The primary contribution is a transparent and reproducible methodology that combines Automated Machine Learning (AutoML) and explainable artificial intelligence (AI) to identify and interpret key risk factors associated with severe crashes.
Using the JADBio AutoML platform, predictive models were constructed to distinguish between severe and non-severe crash outcomes. The models underwent rigorous feature selection across stratified training subsets, and their outputs were interpreted using SHapley Additive exPlanations (SHAP) to quantify the contribution of individual features. A final Ridge Logistic Regression model achieved an AUC-ROC of 85.6\% on the training set and 84.9\% on a hold-out test set, with 17 features consistently identified as the most influential predictors.
Key features spanned demographic, environmental, vehicle, human, and operational categories, including location type, posted speed, minimum occupant age, and pre-crash action. Notably, certain traditionally emphasized factors, such as alcohol or drug impairment, were less influential in the final model compared to environmental and contextual variables. 
Emphasizing methodological rigor and interpretability over mere predictive performance, this study offers a scalable framework to support Vision Zero with aligned interventions and advanced data-informed traffic safety policy.
 \end{abstract}

\begin{graphicalabstract}
\includegraphics[width=\columnwidth]{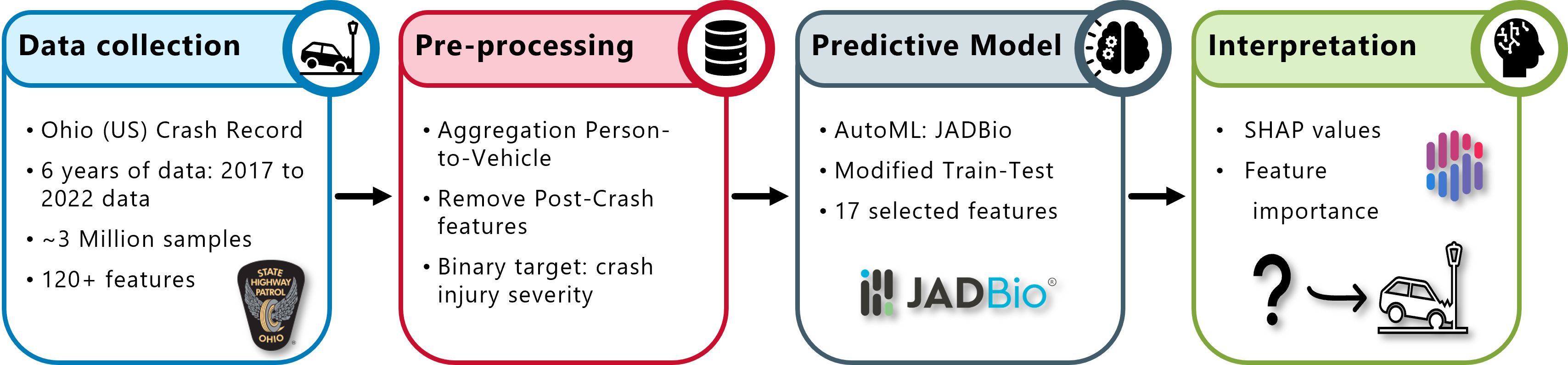}
\end{graphicalabstract}

\begin{highlights}
    \item Introduces a curated six-year dataset of 2.3M vehicle-level crash records from Ohio, US
    \item Demonstrates a reproducible pipeline for large-scale, explainable crash risk modeling
    \item Applies AutoML (JADBio) with SHAP to model and interpret crash severity outcomes
    \item Identifies key predictors across demographic, environmental, and operational domains
    \item Supports Vision Zero-aligned interventions through interpretable, data-driven insights 
\end{highlights}


\end{frontmatter}




\section{Introduction}\label{sec:intro}

Despite efforts to improve road safety, road traffic deaths have increased globally, rising from 1.18 million in 2004 to 1.35 million in 2016 \cite{wang2025global}.
Motor vehicle crashes continue to be a leading cause of death and injury worldwide, imposing significant social and economic burdens on individuals and communities \cite{who, TANDRAYENRAGOOBUR2024}.
Understanding trends in crash data is critical to advancing road safety and reducing the incidence and severity of these events \cite{shannon2018applying}. 
Analyses of both current and retrospective crash data provide valuable insights into underlying causes, including temporal shifts in crash rates, types, and the demographic groups most affected \cite{dong2025evaluation}. 

Such analyses are especially relevant in the context of Vision Zero, a global initiative committed to eliminating traffic-related fatalities and serious injuries through systemic safety improvements \cite{johansson2009vision}. 
Unlike traditional approaches that focus on individual behavior, Vision Zero is grounded in the Safe System Approach (SSA), which emphasizes shared responsibility and calls for improvements across infrastructure design, vehicle technology, speed management, road user behavior, and post-crash care \cite{khan2024advancing}. 
Within this framework, retrospective research plays a critical role in shaping data-informed, adaptive, and sustainable safety strategies. Conventional methods for analyzing crash data often rely on predefined statistical models and require extensive domain expertise to select variables and algorithms \cite{ali2024advances}. 
While valuable, these approaches may be limited in their ability to capture the complex, nonlinear relationships inherent in traffic crash datasets. 
Automated Machine Learning (AutoML) offers a promising solution to these limitations by automating key steps in model development, including variable selection, training, and optimization \cite{he2021automl}. 
AutoML enables the rapid and reproducible analysis of large, high-dimensional datasets, uncovering patterns and interactions that may be overlooked by traditional methods \cite{angarita2021bibliometric}.

This advanced data-driven approach supports the goals of Vision Zero by generating actionable insights into crash risk factors and informing the design of safer transportation systems. 
The present study builds on this potential by aiming to improve understanding of the factors contributing to fatal and suspected serious injury crashes. 
Specifically, the research is structured around three objectives: 
(1) compiling and curating six years (2017–2022) of motor vehicle crash reports from Ohio, yielding a vehicle-level dataset of over 2.3 million records \cite{dryadcrash}; which is, at this time, the largest known crash report dataset available; 
(2) applying AutoML to identify key predictors of crash severity; 
and (3) incorporating explainable AI tools to interpret model outputs and evaluate the influence of individual features. 
To achieve these aims, the study employs JADBio \cite{10.1093/bioinformatics/btad545}, an AutoML platform designed for high-dimensional data analysis. 
JADBio enables robust model development and performs causality-informed feature selection to identify the variables most significantly associated with crash severity, which is defined here as the distinction between non-severe outcomes and those involving suspected serious injuries or fatalities. 
To further enhance interpretability, the study integrates SHapley Additive exPlanations (SHAP) \cite{lundberg2017unified, li2022extracting}, a widely used explainable AI method that quantifies the contribution of each feature to the model's predictions \cite{roussou2025investigation}.

This layered methodology emphasizes transparency, reproducibility, and interpretability, offering a scalable framework for state-level crash analysis. Rather than focusing solely on predictive performance, the study prioritizes methodological rigor and practical insight into crash dynamics. 
In contrast to previous work, often constrained by limited datasets, manual feature engineering, or black-box models \cite{ali2024advances, angarita2021bibliometric}, this research demonstrates how a carefully designed AutoML pipeline can reveal meaningful and generalizable patterns in crash data. 
By aligning machine learning innovation with Vision Zero principles, the study contributes a reproducible, data-driven approach to traffic safety research. The findings support the development of targeted interventions and offer a transferable model for other regions seeking to reduce traffic-related fatalities and serious injuries through evidence-based policy and planning.

The remainder of this paper is organized as follows. 
Section~\ref{sec:related} reviews existing literature on crash severity prediction, highlighting current gaps in data availability, methodological approaches, and model interpretability. 
Section~\ref{sec:data} details the compilation, structure, and characteristics of the curated Ohio crash dataset, emphasizing its scale and relevance. 
Section~\ref{sec:method} describes the proposed methodological pipeline, including data preprocessing, predictive modeling using AutoML, feature selection, and model interpretation through SHAP. 
Section~\ref{sec:results} presents the predictive modeling and interpretability analysis results, offering insights into key factors influencing crash severity. 
Finally, Section~\ref{sec:conclusion} concludes the paper by summarizing key contributions, discussing limitations, and outlining directions for future research.

\section{Related Work}\label{sec:related}
The application of Machine Learning (ML) to crash severity prediction has garnered significant research attention due to its potential to enhance traffic safety and prevent injuries. However, existing studies vary widely in dataset size, feature richness, methodological rigor, and the extent of analysis interpretability, highlighting several persistent research gaps.

A major limitation across the literature is the scarcity of large-scale datasets. Ali et al.~\cite{ali2024advances} categorize crash modeling research into crash occurrence, crash frequency, and injury severity prediction. Their review reveals that most injury severity studies rely on datasets of 10,000 to 100,000 samples, with only a few exceeding one million. Behboudi et al.~\cite{behboudi2024recent} echo these findings, noting that most studies suffer from either small sample sizes or limited feature diversity. Both reviews emphasize the need for larger, more comprehensive datasets and improved model interpretability, gaps directly addressed by the present study.

Automated Machine Learning (AutoML) has emerged as an efficient alternative to manual feature engineering and model tuning. 
Angarita et al.~\cite{angarita2021case} demonstrate the potential of AutoML in crash severity prediction, while their subsequent review \cite{angarita2021bibliometric} evaluates leading frameworks such as AutoGluon, auto-sklearn, and TPOT. 
Although AutoML is shown to perform well in transportation research, the integration of interpretability methods within these workflows remains limited. 
Baykal~\cite{baykal2023accident}, for instance, applied AutoML to a relatively large dataset of 1.6 million U.S. crashes but lacked in-depth analysis of feature importance, underscoring the need to pair predictive accuracy with model transparency.

Another widely recognized challenge is data imbalance, severe crash outcomes are comparatively rare, complicating model training. Fiorentini et al.~\cite{fiorentini2020handling} address this issue using random undersampling to improve predictive performance in datasets of up to 200,000 samples. 
Wen et al.~\cite{wen2021applications} further identify data imbalance, complex feature interactions, and the lack of causal interpretation as major methodological hurdles requiring innovative solutions.

Model interpretability and feature selection have also become essential in crash severity research. 
Several studies have employed SHapley Additive exPlanations (SHAP) to evaluate feature importance \cite{sorum2024identification, sattar2023transparent, zahid2024factors, shao2024injury}.
Dong et al.~\cite{dong2022predicting} and Cheng et al.~\cite{cheng2024crash} use SHAP for both global and local interpretability, yielding useful insights despite small sample sizes. 
Similarly, Sattar et al.~\cite{sattar2023transparent} and Sorum and Pal~\cite{sorum2024identification} identify recurring predictors, such as collision type, contributing circumstances, and vehicle characteristics, while demographic factors tend to play a lesser role. Still, most SHAP-based studies rely on relatively small datasets, limiting the generalizability of their conclusions.

Taken together, the literature reveals several key gaps: limited access to large, feature-rich datasets; underutilization of interpretability tools within AutoML pipelines; challenges related to class imbalance; and a lack of scalable, transparent methodologies. 
This study addresses these shortcomings through the use of a uniquely large dataset exceeding 2 million vehicle-level records, a causality-informed AutoML framework, and comprehensive SHAP-based interpretability analysis.

\section{Comprehensive multi-level dataset of motor vehicle crashes in Ohio, USA}\label{sec:data}

\begin{figure}
    \centering
    \includegraphics[width=0.5\linewidth]{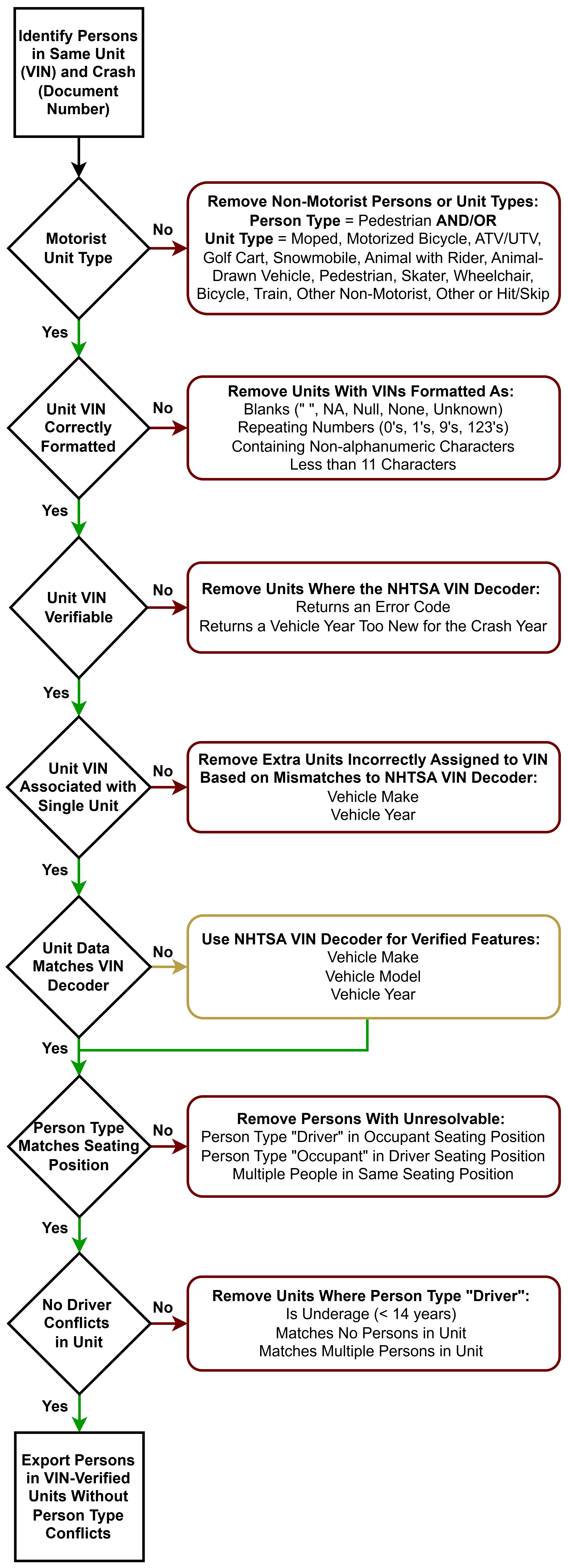}
    \caption{Workflow for verification of unit type, unit VIN, and person type.}
    \label{fig:verification}
\end{figure}

\begin{figure}[tb]
\centering
\begin{minipage}{.38\textwidth}
  \centering
  \includegraphics[width=\linewidth]{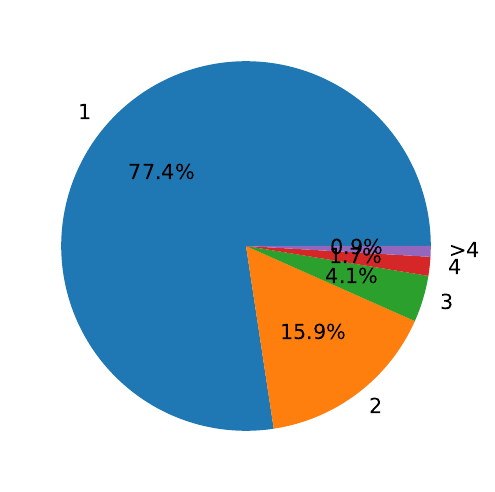}
  \captionof{figure}{Distribution of the number of occupants per vehicle.}
  \label{fig:occupants}
\end{minipage}%
\begin{minipage}{.58\textwidth}
  \centering
  \includegraphics[width=\linewidth]{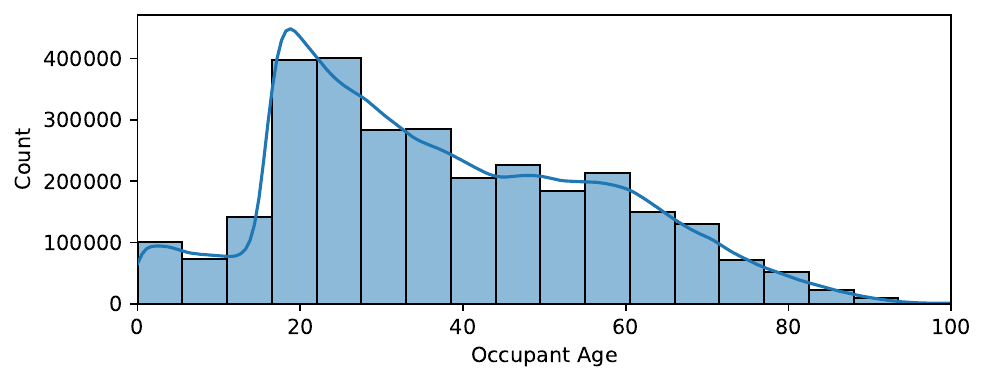} \\
    \includegraphics[width=\linewidth]{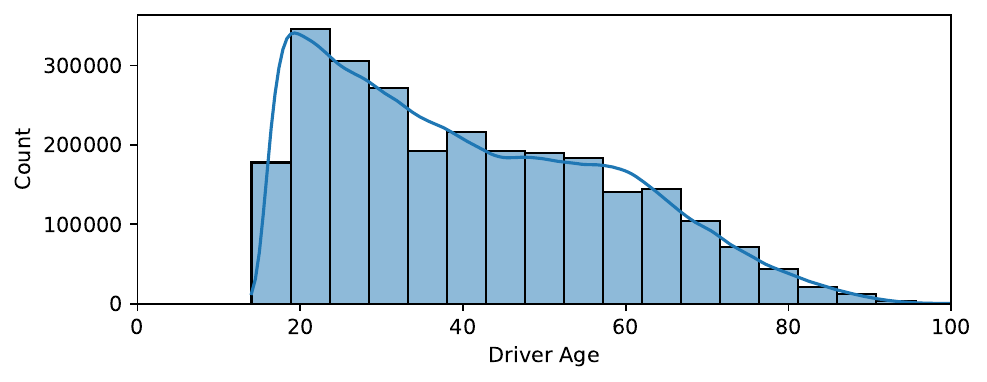}
  \captionof{figure}{Distribution of the occupants (top) and driver age (bottom).}
  \label{fig:ages}
\end{minipage}
\end{figure}

The Ohio Crash Dataset \cite{dryadcrash} is dynamically updated and expanding through the compilation of previous data and currently real-time data collection. 
Traffic crash report data are collected by Ohio law enforcement officers using a standardized crash report form (OH-1) during investigations at the scene of each reported incident. The form captures detailed information about crashes, including vehicle types, environmental conditions, and contributing factors. 
All law enforcement agencies in the state of Ohio submit their Traffic Crash Reports to the Ohio Department of Public Safety (ODPS). These data are collected and then publicly reported through the Ohio Statistics and Analytics for Traffic Safety (OSTATS) system, which standardizes and centralizes the collection of traffic crash reports from across Ohio \cite{ostatsDashboard}. The OSTATS system enables efficient access to comprehensive traffic crash data for analysis and research, ensuring consistency and accuracy in reporting across all jurisdictions within the state. This centralized data collection process provides a robust dataset for identifying trends, conducting statistical analyses, and developing targeted traffic safety initiatives.

 Data were collected at three levels: crash, unit (vehicle), and person (occupant). Each row in the dataset represents a discrete person within a unit involved in the crash. The crash is identified across person rows by a shared CrashID number identifier. Crash-level features report the circumstances of the incident as a whole, such as the date, time, location, road structure, and conditions of the environment and road. Some crash-level features summarize crash outcomes, such as the total number of persons injured or killed and the most severe injury (crash severity). Other crash-level features report the presence or absence of potential risk factors in any unit or person involved in the crash, such as certain unit types (motorcycles, semi-trucks), person age groups (youth, teen, senior), and person behaviors (impairment, speeding).  
 The unit (vehicle) level is defined across person rows by a shared Vehicle Identification Number (VIN). Unit-level features document information about the vehicle (e.g., type, make, model, year) and vehicle movements, actions, and damage. The person (occupant) level reports the person type (driver, occupant, or pedestrian), seating position, demographic information (e.g., age, gender), behavior (e.g., safety equipment, ejection, drug or alcohol impairment, mental condition, distraction), and injury outcome. 
 
 As this dataset is the aggregation of many individual traffic crash reports, it contained some identifiable inaccuracies at the unit (vehicle) and person levels. A data cleaning workflow, depicted in Figure~\ref{fig:verification}, was developed to remove vehicles with unverifiable VINs and persons with uncertain locations inside the vehicle. First, non-motorist person and unit types were excluded. Next, vehicle VINs were validated using the National Highway Traffic Safety Administration (NHTSA) VIN Decoder \cite{NHTSAvindecoder}. All person rows associated with a vehicle were excluded if the VIN was not correctly formatted for verification (e.g., blank, repeating characters, non-numeric characters, too short), returned an error code from the NHTSA VIN Decoder, or returned a vehicle year too recent to have been manufactured before or during the associated crash year. In rare cases, multiple vehicles in a crash were assigned the same VIN. The correct vehicle for that VIN could sometimes be identified by matching the vehicle make and year reported by the NHTSA VIN Decoder to the vehicle make and year in the the traffic crash report. Following this VIN verification, the "verified" vehicle make, model, and year were defined using the NHTSA VIN Decoder values, to eliminate inaccuracies or slight variants manually entered in the traffic crash report. To define unit type, both sources were used, as the traffic crash report provides a more detailed and specific unit type, while the "verified" NHTSA VIN Decoder unit type is a broader category of vehicle. 
 
 Person-level inaccuracies were identifiable as mismatches between the person type (driver or occupant) and seating position within the vehicle. These mismatches were corrected where possible by comparing all persons in the vehicle. For example, the person in the driver's seating position ("front left side") was sometimes incorrectly assigned the person type "occupant". This could be corrected to person type "driver" if no other persons in the vehicle also had the driver person type or seating position. Similarly, units sometimes contained multiple persons assigned person type "driver", only one of whom was in the driver's seating position. Given this confirmed driver, the inaccurately named "drivers" in occupant seating positions could be corrected to person type "occupant". If conflicts between person type and seating position could not be resolved through such comparisons, the uncertain person rows were removed. If the unit's driver had a questionable identity, such as an underage driver (less than 14 years old), no driver, or multiple drivers, the entire unit was removed. Finally, person age was confirmed by subtracting the person's date of birth from the crash date as listed in the traffic crash report. Age category variables were also generated, with children divided into categories of 0-8 years and 9-13 years, and teenagers and adults grouped by five-year or decade categories.   

Following curation, the dataset utilized in this research covers the period from January 1, 2017, to December 31, 2022.
There are in total 119 features in the dataset that characterize each crash in three levels: crash, unit (vehicle), and person (occupant). The complete list of features is found in the \ref{app:original_feature}.
The comprehensive dataset includes detailed records of 1,444,011 crashes, involving a total of 2,280,538 vehicles (units) and 3,069,237 occupants. 
The curated dataset is publicly available \cite{dryadcrash}, and accessible via the unique doi: \url{https://doi.org/10.6084/m9.figshare.29437694}.

Crashes are classified by severity, with the majority (85.0\%) reporting \textit{No Apparent Injury}. \textit{Possible Injury} occurred in 7.1\% of cases, \textit{Suspected Minor Injury} in 6.6\%, \textit{Suspected Serious Injury} in 1.05\%, and \textit{Fatal} accounted for approximately 0.22\% of crashes. 
Each crash involved on average 1.58 ($\pm 0.6$) vehicles, with a maximum of 75 vehicles reported in a single incident. 
Crash frequencies varied annually, with a peak of 435,284 crashes recorded in 2018 and a minimum of 297,639 crashes in 2022. 

The dataset includes details on vehicles involved in the crashes. 
Passenger cars represent the largest proportion of involved vehicle types (47.9\%), followed by multipurpose passenger vehicles (33.1\%), and trucks (14.8\%). A total of 412 different manufacturers and 3,195 unique vehicle models are represented, with Chevrolet, Ford, and Honda being the three most common manufacturers involved.

Occupant demographics and safety characteristics are also documented. 
On average, each vehicle carried 1.35 ($\pm 0.99$) occupants. 
The chart of the number of occupants per vehicle is depicted in Figure~\ref{fig:occupants}.
The dataset includes a nearly balanced distribution of occupant gender, with 45.8\% female and 53.0\% male occupants. 
The mean age of occupants was 37.4 years, while drivers specifically had a mean age of approximately 40.8 years, ranging between 14 and 110 years. Their distribution is reported in Figure~\ref{fig:ages}.

Environmental conditions at the time of the crash were predominantly clear (58.4\%), with cloudy conditions present in 24.8\% of incidents and rain in 11.2\%. 
The majority of crashes occurred during daylight (70.9\%), with other notable lighting conditions including dark but lighted roadways (14.0\%) and dark roadways without lighting (9.3\%). Road conditions were primarily dry (74.8\%), followed by wet (19.8\%).

Key contributing circumstances and operational behaviors leading to crashes are documented. 
The most frequently reported contributing circumstance was following too closely (14.5\%), followed by failure to yield (8.2\%) and other improper actions (7.6\%). 
In terms of pre-crash actions, most vehicles were proceeding straight ahead (56.8\%) or slowing/stopped in traffic (18.5\%). Alcohol and drug involvement were relatively infrequent but notable, reported in 3.3\% and 1.3\% of vehicle incidents, respectively.

This rich and meticulously structured dataset provides a foundational resource for comprehensive analyses at understanding crash dynamics. 
To the best of the authors’ knowledge \cite{ali2024advances}, this represents the largest available dataset of its kind in terms of both sample size and the number of detailed descriptors (features).

\section{Methodology}\label{sec:method}

\begin{figure}[tb]
    \centering
    \includegraphics[width=\linewidth]{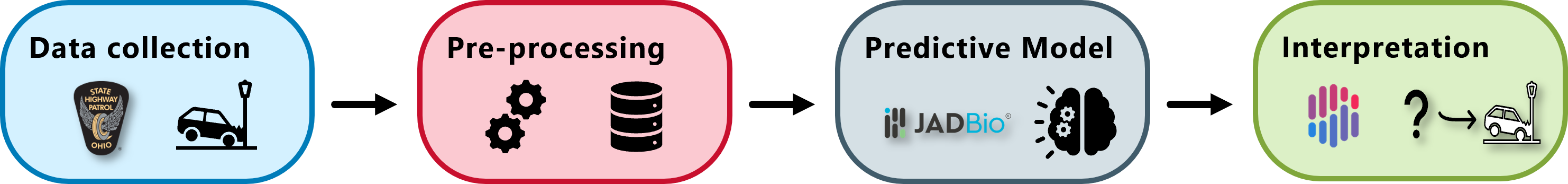}
    \caption{Proposed methodology pipeline.}
    \label{fig:pipeline}
\end{figure}

This section outlines the methodological pipeline used in the study, which includes data collection and preprocessing, predictive modeling using AutoML with feature selection, and model interpretation through SHAP values analysis, as illustrated in Fig.~\ref{fig:pipeline}. 
The raw crash data were first cleaned and aggregated to enable vehicle-level analysis. JADBio, an AutoML platform, was then used to construct and validate predictive models while selecting the most relevant features. Finally, SHAP were used to interpret model outputs and provide transparent insights into the factors influencing crash severity. 

\subsection{Data Pre-processing}\label{sec:preprocessing}
The dataset, described in Section~\ref{sec:data}, consists of approximately 3 million individuals involved in crashes reported by law enforcement, each representing a person involved in a motor vehicle crash. It contains 125 features capturing demographic characteristics, vehicle information, environmental conditions, and pre-crash behaviors. Several preprocessing steps were applied to prepare the data for predictive modeling. 

\paragraph{Data Cleaning} 
Initial cleaning involved removing irrelevant or administrative variables, such as county identifiers. Continuous variables with sporadic missing values (e.g., age, posted speed) were imputed using feature means. For categorical variables, missing values were handled by introducing a "missing" category. 

\paragraph{Post-crash Feature Removal}
To ensure predictive validity, variables reflecting post-crash outcomes (e.g., “most harmful event,” “number of fatalities”) were excluded, ensuring the model relied solely on information available before or at the time of the crash. 

\paragraph{Sample Aggregation} 
Because the dataset was originally structured at the individual level, records were aggregated to the vehicle level by creating the Vehicle Identification Number (VIN). 
Vehicle-specific features were preserved, while occupant-level features were aggregated into summary variables. 
For example, the most severe injury among all occupants defined the target variable, "vehicle severity." Driver-specific variables (e.g., mental condition, distraction, age, and gender) were assigned directly to the vehicle record. Age-related variables were also summarized using meta-features such as minimum, mean, and maximum occupant age.
Information on other vehicles involved in the crash was included by linking up to five additional units using the shared "CrashID." Features such as vehicle model, type, and year of manufacture were extracted and merged accordingly.
 
\paragraph{Data Filtering} 
The analysis focused on passenger and passenger-like vehicles (e.g., passenger cars, SUVs). While motorcycles, trucks, and other vehicle types were not primary subjects, they remained in the dataset as interacting vehicles in crashes involving a passenger vehicle, maintaining the relevance of crash dynamics for vehicle-to-vehicle interactions. 

\paragraph{Data Encoding} 
After aggregation and filtering, the dataset contained 62 features (see~\ref{app:feature}). 
Categorical and cyclical variables were encoded for modeling. Cyclical features such as "YearMonth," "WeekDay," and "DayTime" were encoded using circular transformations to preserve their periodic structure (e.g., 12 months, 7 days, 24 hours). 
Categorical variables were one-hot-encoded into binary features. For instance, the variable "VehicleMake," with 412 unique categories, was expanded into 412 binary columns, each representing the presence of a specific manufacturer. 

\paragraph{Creation of Target Variable} 
Crash severity was originally recorded using five ordinal categories: No Apparent Injury (85.0\%), Possible Injury (7.1\%), Suspected Minor Injury (6.6\%), Suspected Serious Injury (1.0\%), and Fatal (0.2\%). To address the extreme class imbalance, these were collapsed into a binary classification: "Non-Severe" (first three categories) and "Severe" (Suspected Serious Injury and Fatal). The resulting binary target variable retained a class imbalance ratio of approximately 100:1, which, while still challenging, improved modeling feasibility.

Following these preprocessing steps, the final dataset included approximately 2.3 million vehicle-level samples, 4,000 encoded features, and a binary outcome indicating crash severity.

\subsection{Predictive Modeling and Feature Selection} 
The prediction of vehicle crash severity is performed using JADBio \cite{tsamardinos2022just}, an AutoML platform designed specifically for efficient analysis of large-scale, high-dimensional datasets. 
The objective is twofold: first, to build a robust predictive model capable of accurately discriminating severe from non-severe crashes; second, to identify the minimal optimal subset of features necessary to represent the data, hence enhancing interpretability and generating actionable insights.

JADBio provides an extensive search space that includes several modeling algorithms: Decision Trees (DT), Random Forests (RF), Support Vector Machines (SVMs), and Ridge Logistic Regression (Ridge LR), each associated with a comprehensive set of hyperparameters. 
This allows JADBio to systematically explore numerous model configurations to identify the optimal solution in terms of predictive performance.
A complete summary of the model and hyperparameters search space is found in the \ref{app:search_space}.

Given the substantial size of our dataset (approximately 2.3 million samples with around 4,000 features) and the pronounced class imbalance, conventional modeling approaches face computational limitations and risk significant bias towards the majority (non-severe) class.
JADBio addresses class imbalance in several ways: applying stratified CV i.e., each fold retains the class imbalance of the original dataset; adjusting predictive modeling algorithms by e.g. adjusting the cost weights in SVMs; employing the Bootstrap Bias Correction (BBC) method \cite{pmlr-v256-paraschakis24a} that corrects the estimate of the performance of the best model for the “winner’s curse”; having a default threshold-free optimization metric.

Furthermore, for the ease of computational effort, we adopt a modified train-test strategy by constructing four independent training subsets, each with approximately 55,000 samples. 
These subsets are derived via stratified random sampling of the full dataset, preserving the original severe-to-non-severe class ratio. 
This approach guarantees adequate representation of the minority class while maintaining the statistical distribution of the data, thereby facilitating the identification of meaningful predictive patterns without compromising computational feasibility.

For model evaluation and selection within each training subset, JADBio employs an R-repeated, N-incomplete, stratified, K-fold CV (RNK-CV) accompanied by the Early Stopping and Early Dropping heuristics. The hyperparameters R, N and K are dynamically chosen based on the dataset size and class imbalance. 
Dropping enforces a configuration of consistently low performance to be dropped, and the stopping ends the performance estimation process if the performance is not improving anymore, reducing computational costs without sacrificing model robustness.


The predictive performance of the candidate models is evaluated using the Area Under the Receiver Operating Characteristic Curve (AUC-ROC), and to calculate their Confidence Interval (CI), we use the BBC algorithm \cite{pmlr-v256-paraschakis24a}.
The AUC-ROC, representing the relationship between True Positive Rate (TPR) and False Positive Rate (FPR) across varying classification thresholds provides a threshold-independent measure of model performance, capturing both sensitivity and specificity without being severely influenced by class distribution biases.

Feature Selection (FS) constitutes a critical aspect of our modeling approach. 
Within each training iteration, FS algorithms identify a minimal yet informative subset of predictive features, improving interpretability and reducing computational overhead. 
The FS methods evaluated in our analyses include Epilogi \cite{10.1093/bioinformatics/btad545}, Least Absolute Shrinkage and Selection Operator (LASSO), Univariate Feature Selection with Benjamini-Hochberg correction, and Statistically Equivalent Signatures (SES) \cite{lagani2016feature}. 
The total search space of model configurations amounts to 738, comprising 737 algorithm-FS-hyperparameter combinations and one naive baseline model.
A summary of the specification for the AutoML pipeline is outlined in Table~\ref{tab:automl}.

\begin{table}[tb]
    \centering
    \small
    \begin{tabular}{l l l}
        \toprule
         \textbf{Specification} & \textbf{Value} & \textbf{Comment} \\
         \midrule
         \textit{Predictive task} & Binary Classification & Minor vs Severe (100:1)\\
         \textit{Dataset size} & 2,300,000 x 4,000 & Dataset size after pre-processing\\
         \midrule
         \textit{Repeats} & 4 & Stratified random sampling \\
         \textit{Subset data size} & 55,000 x 4,000 & Dataset size each repetition \\
         \textit{Cross Validation} & 10-fold RNK-CV & CV strategy per each repetition \\
         \textit{Model search space} & 738 configuration / fold & Naive model is added each fold \\
         \textit{Performance metric} & ROC-AUC & Threshold-free metric\\
         \bottomrule
    \end{tabular}
    \caption{Summary of specification for the AutoML pipeline.}
    \label{tab:automl}
\end{table}

To construct a robust and interpretable predictive model, results from the four independent training iterations are aggregated based on the \textit{stability} of selected features. 
The feature stability is defined as the frequency with which a given feature is selected across the different training subsets. 
Using a predefined stability threshold, we retain features consistently identified as predictive across multiple training subsets, as these indicate higher reliability and predictive relevance. 
Specifically, we adopt a stability threshold of 75\%, meaning only features selected in at least three out of four training iterations are included in the final model. 
After selecting the most stable feature subset, a final predictive model is trained on the combined data from all the four subset, and then tested on the hold-out set composed of all the data samples unseen during training.

All experiments were conducted on a workstation equipped with an Intel(R) Xeon(R) Platinum 8272CL CPU @ 2.60GHz/core (5 cores used) and 64 GB RAM. The system operated on Azure Linux 2.0.

\subsection{Model Interpretation}\label{sec:shap}
To enhance the interpretability of the final predictive model, SHapley Additive exPlanations (SHAP)~\cite{lundberg2017unified} were employed, a widely-adopted framework for interpreting complex machine learning predictions \cite{roussou2025investigation}. 
SHAP values provide a unified method to quantify the contribution of each feature to the predictions made by the model, allowing the interpretation of both the magnitude and direction of these impacts on an instance-by-instance basis.

Given a predictive model \( f \) trained on a dataset consisting of \( p \) features, the SHAP framework decomposes the model's prediction for a specific instance \( x = [x_1, x_2, \dots, x_p] \) into a sum of feature-specific contributions:
\begin{equation}
    f(x) = \phi_0 + \sum_{i=1}^{p} \phi_i \, .
\end{equation}
In this formulation, \(\phi_0\) represents the average prediction across the entire dataset, while each \(\phi_i\) indicates how much the \(i\)-th feature deviates from the prediction for instance \( x \) from this baseline. 
Positive SHAP values (\(\phi_i > 0\)) indicate features pushing the prediction towards the positive class, which corresponds to severe injury outcome in our binary problem setting.
Whereas negative SHAP values (\(\phi_i < 0\)) represent features associated with the negative class, hence non-severe outcomes.


\section{Results and Discussion}\label{sec:results}
This section presents the results of the predictive modeling and feature interpretation process. 
It begins with an overview of model development and the feature selection strategy employed to identify the most relevant predictors of crash severity. 
Subsequently, the outputs of the final model are interpreted using SHAP to assess the relative influence of each feature. 

\subsection{Predictive Model and Feature Selection}\label{sec:model_FS}

\begin{table}[tb]
    \centering
    \scriptsize
    \begin{tabular}{ccccccc}
    \toprule
    \textbf{\#} & \textbf{Trained Models} & \textbf{Runtime} & \textbf{Features} & \textbf{FS} & \textbf{Model} & \textbf{ROC-AUC} \\
    \midrule
    1 & 6,633 & 214 h & 62   &\makecell{SES \\ ($a=0.05$)} & \makecell{Ridge LR \\ ($\lambda=1)$} & \makecell{83.4\% \\ $[80.6\%-86.0\%]$}\\
    2 & 6,633 & 140 h & 65   & \makecell{SES \\ ($a=0.05$)} & \makecell{Ridge LR \\ ($\lambda=1)$} & \makecell{84.9\% \\$[ 82.5\% - 87.3 \% ]$} \\
    3 & 6,633 & 187 h & 100  & \makecell{SES \\ ($a=0.1$)} & \makecell{Ridge LR \\ ($\lambda=10)$} & \makecell{83.3\% \\$[ 80.4\% - 86.0 \% ]$}\\
    4 & 6,633 & 154 h & 22   & \makecell{SES \\ ($a=0.01$)} & \makecell{RF \\ ($nT=1000$)} & \makecell{85.3\%\\ $[ 82.8\% - 87.6 \% ]$} \\
    \bottomrule
    \end{tabular}
    \caption{Performance of the modified Train-Test approach, and the selected best model per train iteration (\#) by the AutoML pipeline.}
    \label{tab:train_test_split}
\end{table}

As described in Section~\ref{sec:method}, following data pre-processing, four disjoint training subsets were constructed, each containing approximately 55,000 samples, to support robust identification of an optimal minimal feature set for crash severity prediction. 
The outcomes of each training iteration are summarized in Table~\ref{tab:train_test_split}. 
A total of 6,633 models were trained per iteration, based on the 10-fold RNK-CV strategy. 
Each iteration terminated after nine folds, as no additional statistically significant performance improvements were observed (737 model configurations per fold $\times$ 9 folds = 6,633 models).

Although this modified training and CV approach improved efficiency, computational demands remained high due to the dataset’s size and dimensionality. On average, each training iteration required approximately 174 hours of runtime.

Across all four iterations, the SES \cite{lagani2016feature} feature selection algorithm consistently emerged as the most effective, although the number of selected features varied considerably, ranging from 22 to 100.

In terms of predictive performance, Ridge Logistic Regression (Ridge LR) was the top-performing model in three of the four iterations, with ROC-AUC values ranging from 83.3\% to 84.9\%. 
The highest overall performance, however, was achieved using a Random Forest (RF) model, which obtained a ROC-AUC of 85.3\% (95\% CI: 82.8\%–87.6\%) when paired with a more restrictive SES threshold ($\alpha = 0.01$), selecting only 22 features.
Importantly, performance across all four iterations was comparable, as reflected in overlapping 95\% confidence intervals ranging from 80.6\% to 87.6\%. This consistency suggests that, despite each subset representing a different segment of the data, they captured similar underlying patterns and predictive signals.

To derive the final feature set, the features identified by SES in each training subset were aggregated and evaluated for selection stability, defined as the frequency with which a feature appeared across the four training iterations. 
A threshold of 75\% was applied, retaining features selected in at least three out of four subsets. 
This process yielded a final set of 17 stable features, presented in Table~\ref{tab:categories}.
The final feature signature includes a mix of numerical and categorical variables. To facilitate interpretation and downstream analysis, these features were grouped into five thematic categories: Demographic, Human, Environmental, Vehicle, and Operational; following a classification scheme similar to that proposed by Ali et al.~\cite{ali2024advances}.

\begin{table}[tb]
    \centering
    \small
    \begin{tabular}{l l c c}
    \toprule
    \textbf{Category} & \textbf{Feature} & \textbf{Occurrence} & \textbf{Data Type}  \\
    \midrule
    \multirow{3}{*}{Demographic} & Driver Age & 3/4 & Numerical \\
    & Occupant(s) Mean Age & 3/4 & Numerical \\
    & Occupant(s) Minimum Age & 4/4 & Numerical \\
    \midrule
    \multirow{6}{*}{Human} & Driver Condition & 4/4 & Categorical \\
    & Driver Distraction & 4/4&  Categorical \\
    & Seat Belt Status (Belted) & 4/4 & Categorical \\
    & Posted Speed* & 4/4 & Numerical \\
    & Alcohol Impairment & 3/4 & Categorical \\
    & Drug Impairment & 4/4 & Categorical \\
    \midrule
    \multirow{3}{*}{Environmental} & Road Contour & 3/4 & Categorical \\
    & Location & 4/4 & Categorical\\
    & Animal Related  & 4/4 & Categorical \\
    \midrule
    \multirow{3}{*}{Vehicle} & Interacting Vehicle Type & 3/4 & Categorical \\
    & Vehicle Year & 4/4 & Numerical \\
    & Number of Occupants & 3/4 & Numerical \\
    \midrule
    \multirow{2}{*}{Operational} & Contributing Circumstance & 4/4 & Categorical \\
    & Pre-Crash Action & 4/4 & Categorical \\
    \bottomrule
    \multicolumn{4}{l}{\footnotesize*: Posted Speed is used as an indicator for the speed of the car.}
    \end{tabular}
    \caption{Final Signature Set Features and Categories with their feature stability.}
    \label{tab:categories}
\end{table}

\begin{figure}[tb]
    \centering
    \includegraphics[width=0.87\linewidth]{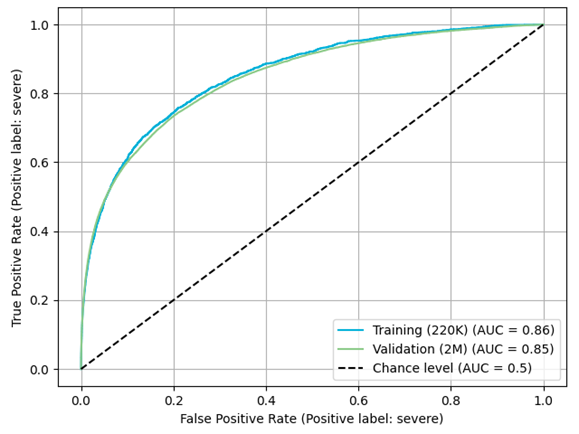}
    \caption{ROC-AUC plots for the final model. Training performance is calculated on the union of samples participating in the 4 sub-analyses, when training a RidgeRegression model on the features present in at least 3 out of 4 sub-analyses, and using 220,000 samples.}
    \label{fig:roc}
\end{figure}

After finalizing the optimal subset of 17 high-stability features, the four previously constructed training subsets were aggregated into a single training set comprising 220,000 samples. 
The remaining approximately 2 million vehicle-level records, which had not been used in any prior modeling steps, were reserved as an independent hold-out test set.
A Ridge LR model with a regularization parameter of $\lambda = 1$ was trained using the aggregated training set. 
Ridge LR was selected as the final model over RF due to its comparable predictive performance, improved interpretability, and lower computational complexity.
As illustrated in Figure~\ref{fig:roc}, the final Ridge LR model achieved a ROC-AUC of 85.58\% on the training data and 84.91\% on the hold-out test set (CI: 84.7\%-86.4\%). The close alignment between training and testing performance demonstrates strong generalization capabilities, with no evidence of overfitting or underfitting.
It is also noteworthy that the model's performance remains comparable to that achieved previously using the complete feature set, from Table~\ref{tab:train_test_split}, despite now utilizing only 17 selected features.

\subsection{Feature Interpretation using SHAP}

\begin{figure}[tb]
    \centering
    \includegraphics[width=0.97\linewidth]{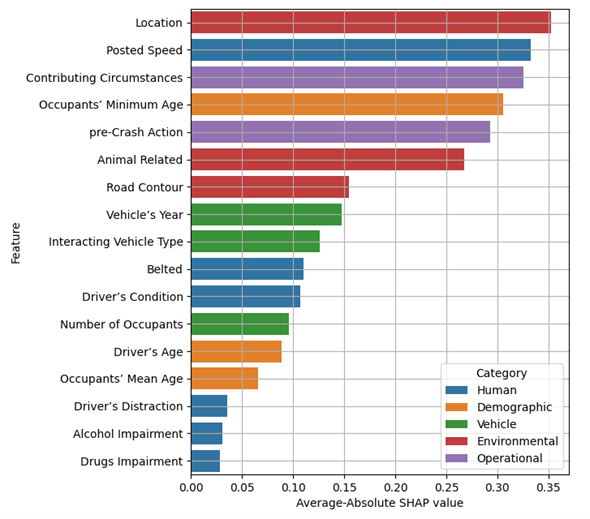}
    \caption{The final feature set (stability >75\%), sorted by feature importance (average absolute SHAP values).}
    \label{fig:feature_importance}
\end{figure}

To interpret and explain the predictions of the final model, SHAP values were employed. 
Although the selected model, Ridge LR, provides inherent interpretability through its linear coefficients, SHAP offers instance-level explanations that quantify the direction and magnitude of each feature’s contribution to individual predictions \cite{lundberg2017unified}. 
This capability is particularly valuable in high-dimensional datasets, where complex feature interactions and non-linear effects may still influence predictions, even in linear models.
Moreover, in linear models with independent features, SHAP values can directly correspond to regression coefficients, providing a clear and consistent global interpretation while retaining flexibility for local explanations \cite{christoph2020interpretable}.

To quantify feature importance, SHAP values were computed for each instance in the dataset. The overall importance of each feature was then calculated by averaging the absolute SHAP values across all samples:
\begin{equation}
    VI_j = \frac{1}{N}\sum_{i=1}^{N}|S_{ij}| \,,
\end{equation}
where \(VI_j\) represents the variable importance for feature \(j\), \(S_{ij}\) is the SHAP value of feature \(j\) for sample \(i\), and \(N\) denotes the total number of samples. 
These average importance scores were subsequently used to rank features according to their influence on predicted crash severity outcomes.

The ranked feature importance results are presented in Figure~\ref{fig:feature_importance}, with each feature color-coded by category (as defined in Section~\ref{sec:model_FS}). 
The most influential predictor is \textit{Location}, an environmental variable that distinguishes among contexts such as “city,” “township,” and “village,” reflecting the substantial variability in crash severity across urban and non-urban areas. 
\textit{Posted Speed} and \textit{Contributing Circumstances} follow as highly influential, underscoring the impact of operational speed limits and pre-crash situational factors on injury severity.

Among demographic variables, \textit{Occupants' Minimum Age} emerged as especially important, indicating that the presence of younger passengers is a significant determinant of crash severity, more so than \textit{Driver Age}, which ranked lower in influence. Operational factors such as \textit{Pre-Crash Action} also demonstrated strong predictive value, reinforcing the importance of behaviors and decisions made immediately before a collision.

In the vehicle category, \textit{Vehicle Year} and \textit{Interacting Vehicle Type} were identified as key predictors. 
They highlight the relevance of vehicle safety features, which are often associated with newer models, and the type of vehicle involved in the crash (e.g., passenger car vs. truck), which may affect injury mechanisms and severity.

Interestingly, features commonly assumed to be high-risk factors \cite{safari2020comprehensive}, such as \textit{Alcohol Impairment}, \textit{Drug Impairment}, and \textit{Driver Distraction}, exhibited comparatively lower importance in the model. 
This may be due to their lower prevalence within the dataset or the stronger relative influence of environmental and contextual variables.
It is also noteworthy that the top ten most influential features span all five defined categories (Demographic, Human, Environmental, Vehicle, and Operational), underscoring the complex and multifactorial nature of crash severity prediction.

\begin{figure}[tb]
    \centering
    \includegraphics[width=0.95\linewidth]{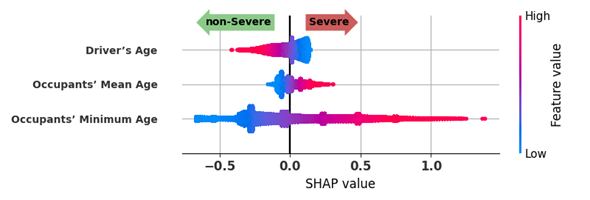}
    \caption{SHAP values corresponding to demographics category of identified features.}
    \label{fig:demographics}
\end{figure}

\begin{figure}[tb]
    \centering
    \includegraphics[width=0.95\linewidth]{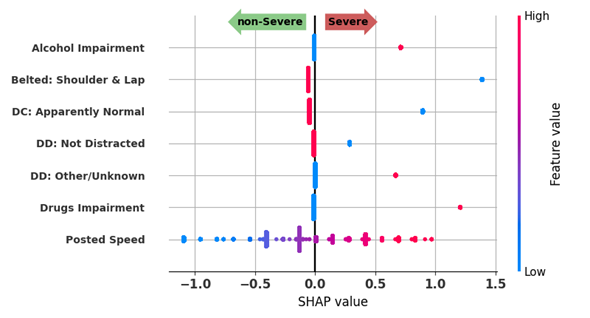}
    \caption{SHAP values corresponding to human factors category of identified features.}
    \label{fig:human}
\end{figure}

\begin{figure}[tb]
    \centering
    \includegraphics[width=0.95\linewidth]{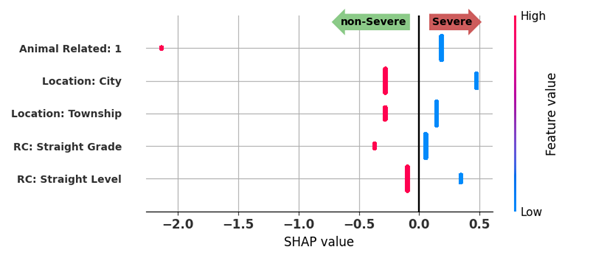}
    \caption{SHAP values corresponding to the environmental factors category of identified features.}
    \label{fig:environmental}
\end{figure}

\begin{figure}[tb]
    \centering
    \includegraphics[width=0.95\linewidth]{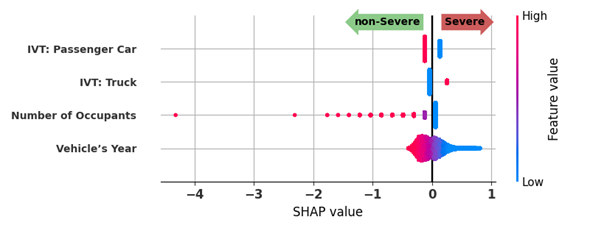}
    \caption{SHAP values corresponding to the vehicle characteristics category of identified features.}
    \label{fig:vehicle}
\end{figure}

\begin{figure}[tb]
    \centering
    \includegraphics[width=0.95\linewidth]{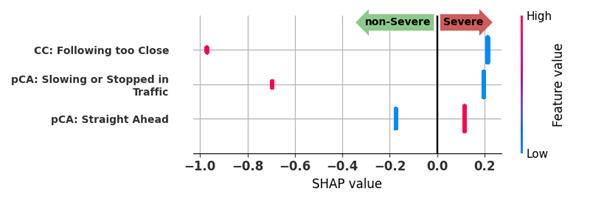}
    \caption{SHAP values corresponding to the operational category of identified features.}
    \label{fig:operational}
\end{figure}

To evaluate the effect of each feature and its individual levels on crash severity predictions, SHAP summary plots were generated (Figures~\ref{fig:demographics}--\ref{fig:operational}). 
These visualizations show how each feature affects the predicted likelihood of the two classes: "severe outcomes" and "non-severe outcomes". 
They indicate both how strongly and in which direction each feature influences the prediction.
In each plot, SHAP values are centered at zero: negative values indicate a decrease in predicted severity (i.e., a non-severe outcome), while positive values indicate an increased risk of a severe outcome, as explained in Section~\ref{sec:shap}.
Because SHAP values are expressed in absolute terms, they are directly comparable across features and analyses.
The data points in the SHAP plots are color-coded on a gradient from blue (representing low feature values) to red (representing high feature values). 
For categorical features, one-hot encoding expands each variable into multiple binary indicators representing individual levels. 
Due this encoding strategy, to assess the overall importance pf the categorical features, the maximum average absolute SHAP value across its levels was used. 
To enhance interpretability and reduce visual complexity, only categorical levels with importance values exceeding 40\% of the most influential level within the same feature were retained in the visualizations. 
A complete analysis of all 17 features and their categorical levels is provided in the~\ref{app:shap_full}.

Figure~\ref{fig:demographics} illustrates the distribution of SHAP values for the three continuous demographic features: \textit{Driver’s Age}, \textit{Occupants’ Mean Age}, and \textit{Occupants’ Minimum Age}. 
Consistent with the feature importance rankings shown in Figure~\ref{fig:feature_importance}, \textit{Occupants’ Minimum Age} has the widest range of SHAP values, approximately from -0.7 to 1.4, suggesting it strongly influences the predicted severity of crashes. 
Younger vehicle occupants (indicated by blue colors and lower values) have negative SHAP scores, signifying a lower risk of severe injury, while older occupants (indicated by red colors and higher values) have positive SHAP scores, pointing to a higher risk of severe injury. 
Thus, the age of the youngest occupant in a vehicle is particularly important for predicting crash severity, with younger occupants typically experiencing less severe outcomes. A similar, although less significant, trend appears for \textit{Occupants’ Mean Age}.  These findings are consistent with existing literature indicating that older individuals have an increased risk of severe injuries due to physiological factors and lower trauma tolerance~\cite{newgard2008defining}.

In contrast, the \textit{Driver’s Age} reveals a more complex relationship with crash severity. 
Younger drivers (represented in blue) are associated with higher injury severity predictions, whereas older drivers (represented in red) tend to correlate with lower severity. 
This observation supports prior literature \cite{ryan1998age, de2010driver}, as this pattern can be interpreted that younger drivers may exhibit more reckless and risking driving behaviour, leading to severe crash outcomes.
Recall that our analysis is purely data-driven, without incorporating explicit domain knowledge.

Figure~\ref{fig:human} visualizes the SHAP values for human-related features. 
Within this category, the most influential feature is \textit{"Posted Speed"}, exhibiting a symmetrical relationship, with higher speeds associated with increased crash severity, and lower speeds associated with non-severe outcomes.
Features such as \textit{"Alcohol Impairment"} and \textit{"Drugs Impairment"} strongly increase predicted severity (positive SHAP values).
However, these occurrences are infrequent, as indicated by a few scattered red points to the right compared to the more numerous blue points clustered near zero. 
Conversely, consistent seatbelt usage (\textit{"Belted: Shoulder \& Lap"}) and normal driver conditions (\textit{"DC: Apparently Normal"}) predominantly yield negative SHAP values, reflecting their protective role in reducing crash injury severity. 
Furthermore, the absence of seatbelt usage (blue points in the \textit{"Belted: Shoulder \& Lap"} feature) is strongly correlated with severe outcomes, aligning well with established safety literature that emphasizes the critical role of seat belts in injury prevention~\cite{febres2020influence}.

The environmental-related features are analyzed in Figure~\ref{fig:environmental}.
The feature \textit{"Animal Related: 1"} has a strongly negative SHAP value, indicating that crashes involving animals are generally associated with lower severity outcomes. 
Regarding location, \textit{"Location: City"} and \textit{"Location: Township"} exhibit mostly negative SHAP values, suggesting that crashes within urbanized areas (higher feature values, red points) are typically associated with non-severe outcomes. 
In contrast, crashes outside urbanized areas (lower feature values, blue points) tend toward higher severity risk.
The road contour features (\textit{"RC: Straight Grade"} and \textit{"RC: Straight Level"}) show mixed associations, indicating complex interactions between road geometry and crash severity, most likely depending on other contextual factors.
Overall, this specific analysis highlights the protective role of urban environments on crash severity, and underscores the complex influence of road conditions.

Figure~\ref{fig:vehicle} illustrates the influence of vehicle-related features on crash severity prediction.
\textit{"Vehicle’s Year"} is the most important feature of this group.
It shows that older vehicles (blue points) slightly increase severity risk, while newer vehicles (red points) tend toward reduced severity, likely reflecting improvements in vehicle safety over time \cite{furlan2020advanced}. 

The feature \textit{"IVT: Passenger Car"} (Interacting Vehicle Type: Passenger Car) predominantly has negative SHAP values, indicating collisions involving passenger cars only, typically result in less severe outcomes, potentially due to differences in driver demographics and relationships with human factors.
The feature \textit{"IVT: Truck"} exhibits positive SHAP values, suggesting collisions involving trucks and passenger cars generally increase the injury severity for the passenger car.
Interestingly, the \textit{"Number of Occupants"} shows that vehicles with higher numbers of occupants (red points) are slightly associated with reduced severity, possibly because additional passengers encourage more cautious driving behavior.

Figure~\ref{fig:operational} presents the SHAP values for the operational category.
Although both categories, \textit{"Contributing Circumstance" (CC)} and \textit{"Pre-Crash Action" (pCA)},  contain multiple levels, the plot highlights the influence of specific conditions. 
\textit{"Following too Close"} exhibits a strong negative SHAP value, potentially due to the frequency of this variable in all levels of vehicle severity. 
The same effect is demonstrated for \textit{"Slowing or Stopped in Traffic"}, as given the reduced speed, it is most likely to incur non-severe crashes.
Conversely, the pre-crash action \textit{"Straight Ahead"} suggests a higher probability of a severe crash when this action is related to it. 
This counter-intuitive result might be explained by the potential for elevated speeds associated with traveling straight, thereby increasing the severity of crashes that do occur.

\section{Conclusions}\label{sec:conclusion}
This study contributes to the Vision Zero mission by introducing a transparent, reproducible, and data-driven framework for predicting crash severity using Automated Machine Learning (AutoML) and explainable AI (xAI). Through the curation and public release of over 2.3 million vehicle-level crash records from Ohio (2017–2022), the authors provide one of the most comprehensive open-access crash datasets currently available. This resource offers critical infrastructure for research replication, cross-jurisdictional comparison, and policy evaluation.

The study presents a novel methodological pipeline that prioritizes interpretability without sacrificing predictive performance. By applying causality-informed feature selection and SHAP-based explanation methods, the model distilled over 4,000 candidate features into a robust and stable subset of 17 predictors. The final Ridge Logistic Regression model achieved a ROC-AUC of 84.9\% on an independent test set, demonstrating generalizability across millions of real-world crash records.

Key findings highlight the influential roles of environmental, demographic, and operational factors, such as location type, posted speed, and occupants' minimum age, in determining crash severity. 
In contrast, traditionally emphasized behavioral indicators like alcohol or drug impairment showed lower predictive importance. These insights reinforce the Safe System Approach (SSA), which shifts the focus from individual fault to systemic risk, emphasizing road design, speed management, and vehicle safety as levers for injury prevention.
Notably, our purely data-driven analysis, conducted without incorporating explicit domain knowledge, aligns closely with findings from established research. 
This consistency confirms the robustness and validity of our methodological approach.

In comparison to many prior studies that rely on smaller datasets or black-box models, this work demonstrates how interpretable machine learning can support evidence-based decision-making. 
The resulting framework enables domain experts, policymakers, and practitioners to identify high-impact variables and design data-informed interventions, for example, targeting specific road environments, vulnerable age groups, or pre-crash conditions with context-specific countermeasures.

Despite these advances, the study acknowledges two primary limitations. First, the use of a linear model, though beneficial for transparency, may limit the capacity to capture complex non-linear interactions. 
Second, the extreme class imbalance (approximately 100:1) between non-severe and severe outcomes was not directly addressed via resampling or cost-sensitive learning, which may reduce sensitivity to rare but critical events.

Future work should explore hybrid and interpretable ensemble models that retain explainability while capturing non-linearity. 
Addressing class imbalance through synthetic oversampling or customized loss functions may enhance the detection of severe crashes. 
Finally, integrating causal inference frameworks could elucidate mechanisms of injury severity and further strengthen the translation of model outputs into actionable public safety strategies.

\section*{Acknowledgments} 
The authors gratefully acknowledge the Ohio State Highway Patrol and the Ohio Department of Public Safety for their role in maintaining and providing access to the Ohio Traffic Crash Report data. 
We also extend our sincere thanks to the many law enforcement agencies across Ohio for their efforts in submitting and cataloging these reports. Their continued commitment to accurate data collection and centralization made this research possible.

\noindent This work was partially funded by the Honda Research Institute Europe. 

\section*{CRediT authorship contribution statement}

\textbf{Conceptualization:} Andrea Castellani, Zacharias Papadovasilakis, Giorgos Papoutsoglou, Ioannis Tsamardinos;
b\textbf{Data Curation:} Andrea Castellani, Zacharias Papadovasilakis, Brian Bautsch, Mary Cole, Angela Harden;
rian\textbf{Formal analysis:} Andrea Castellani, Zacharias Papadovasilakis;
\textbf{Funding acquisition:} Andrea Castellani, Brian Bautsch, Tobias Rodemann;
\textbf{Investigation:} Andrea Castellani, Zacharias Papadovasilakis, Giorgos Papoutsoglou;
\textbf{Methodology:} Andrea Castellani, Zacharias Papadovasilakis, Giorgos Papoutsoglou;
\textbf{Project administration:} Andrea Castellani, Tobias Rodemann, Ioannis Tsamardinos;
\textbf{Resources:} Andrea Castellani, Brian Bautsch, Tobias Rodemann;
\textbf{Software:} Andrea Castellani, Zacharias Papadovasilakis;
\textbf{Supervision:} Andrea Castellani, Ioannis Tsamardinos, Angela Harden;
\textbf{Validation:} Andrea Castellani, Zacharias Papadovasilakis, Giorgos Papoutsoglou;
\textbf{Visualization:} Andrea Castellani, Zacharias Papadovasilakis;
\textbf{Writing – original draft:} Andrea Castellani, Zacharias Papadovasilakis, Giorgos Papoutsoglou, Mary Cole, Angela Harden;
\textbf{Writing – review \& editing:} Andrea Castellani, Zacharias Papadovasilakis, Giorgos Papoutsoglou, Mary Cole, Tobias Rodemann, Ioannis Tsamardinos, Angela Harden.

\section*{Declaration of competing interest}
Honda Research Institute Europe and American Honda Motor embrace the Safe System Approach towards reducing road traffic collision fatalities.

\section*{Data Availability}
Data published in \cite{dryadcrash}, and freely available at: \url{https://doi.org/10.6084/m9.figshare.29437694}.
The code for analysis will be made available on request.

\clearpage
\appendix
\section{Original Dataset Features}\label{app:original_feature}
In Table~\ref{tab:full_data_features} are reported the complete list of the 119 features available in the dataset, sorted in alphabetical order.

{
\footnotesize
\begin{longtable}{lllll}
\toprule
                                      \textbf{Level} & \textbf{Feature Name} & \textbf{Type} & \textbf{Representation} \\
\midrule
\endfirsthead

\toprule
                                      \textbf{Level} & \textbf{Feature Name} & \textbf{Type} & \textbf{Representation} \\
\midrule
\endhead
\midrule
\multicolumn{4}{r}{{Continued on next page...}} \\
\midrule
\endfoot

\bottomrule
\endlastfoot
crash &                    ActiveSchoolZoneRelated &      binary &        boolean \\
   crash &                              AnimalRelated & categorical &        integer \\
   crash &                                     Belted & categorical &         string \\
   crash &                                 Crash.Year &   numerical &        integer \\
   crash &                              CrashDateTime &    datetime &        integer \\
   crash &                    CrashLocationInWorkZone & categorical &        integer \\
   crash &                              CrashSeverity & categorical &        integer \\
   crash &                 DividedLaneTravelDirection & categorical &        integer \\
   crash &                             CrashID &       index &        integer \\
   crash &                                    HitSkip & categorical &        integer \\
   crash &                      InCityVillageTownship & categorical &         string \\
   crash &              IntersectionOrApproachRelated &      binary &        boolean \\
   crash &                           IsAlcoholRelated &      binary &        boolean \\
   crash &                           IsBicycleRelated &      binary &        boolean \\
   crash &                        IsCommercialAtFault &      binary &        boolean \\
   crash &                        IsCommercialRelated &      binary &        boolean \\
   crash &                              IsDrugRelated &      binary &        boolean \\
   crash &                             IsDUI21Related &      binary &        boolean \\
   crash &                       IsFatalNotReportable &    constant &        boolean \\
   crash &                        IsMotorcycleRelated &      binary &        boolean \\
   crash &                        IsPedestrianRelated &      binary &        boolean \\
   crash &                         IsSemiTruckRelated &      binary &        boolean \\
   crash &                            IsSeniorRelated &      binary &        boolean \\
   crash &                        IsSmallTruckRelated &      binary &        boolean \\
   crash &                             IsSpeedRelated &      binary &        boolean \\
   crash &                              IsTeenRelated &      binary &        boolean \\
   crash &                             IsYouthRelated &      binary &        boolean \\
   crash &                             LightCondition & categorical &        integer \\
   crash &                  LocationFirstHarmfulEvent & categorical &        integer \\
   crash &                           LocationRoadType & categorical &         string \\
   crash &                          LocationRouteType & categorical &         string \\
   crash &                          MannerOfCollision & categorical &         string \\
   crash &                              NumberOfUnits &   numerical &        integer \\
   crash &                            PrivateProperty &      binary &        boolean \\
   crash &                              RoadCondition & categorical &        integer \\
   crash &                                RoadContour & categorical &        integer \\
   crash &                                RoadSurface & categorical &        integer \\
   crash &                             RoadwayDivided &      binary &        boolean \\
   crash &                             SecondaryCrash &      binary &        boolean \\
   crash &                               TotalInjured &   numerical &        integer \\
   crash &                                TotalKilled &   numerical &        integer \\
   crash &                                UnitInError &      binary &        boolean \\
 vehicle &                                    Units.. &   numerical &        integer \\
 vehicle &                               Units.Action & categorical &        integer \\
 vehicle &                        Units.CargoBodyType & categorical &        integer \\
 vehicle &                        Units.CommercialUse &      binary &        boolean \\
 vehicle &             Units.ContributingCircumstance & categorical &        integer \\
 vehicle &                         Units.DamagedAreas & multi-label &        integer \\
 vehicle &                          Units.DamageScale & categorical &        integer \\
 vehicle &                        Units.DirectionFrom & categorical &        integer \\
 vehicle &                          Units.DirectionTo & categorical &        integer \\
 vehicle &                    Units.FirstHarmfulEvent & categorical &        integer \\
 vehicle &                        Units.GovernmentUse &      binary &        boolean \\
 vehicle &                         Units.HasHMPlacard &      binary &        boolean \\
 vehicle &            Units.HazardousMaterialReleased &      binary &        boolean \\
 vehicle &                          Units.HitSkipUnit &      binary &        boolean \\
 vehicle &                        Units.HMClassNumber & categorical &        integer \\
 vehicle &                      Units.HMPlacardNumber &       index &        integer \\
 vehicle &                  Units.InEmergencyResponse &      binary &        boolean \\
 vehicle &                Units.InitialPointOfContact & categorical &        integer \\
 vehicle &              Units.InterlockDeviceEquipped &      binary &        boolean \\
 vehicle &                       Units.IsMotoristUnit &      binary &        boolean \\
 vehicle &                     Units.IsNonContactUnit &      binary &        boolean \\
 vehicle &                    Units.IsNonMotoristUnit &      binary &        boolean \\
 vehicle &                          Units.IsTrainUnit &      binary &        boolean \\
 vehicle &                     Units.MostHarmfulEvent & categorical &        integer \\
 vehicle &          Units.NonMotoristLocationAtImpact & categorical &        integer \\
 vehicle &                    Units.NumberOfOccupants &   numerical &        integer \\
 vehicle &                    Units.NumberOfThruLanes &   numerical &        integer \\
occupant &                             Units.People.. &   numerical &        integer \\
occupant &                Units.People.Age.Calculated &   numerical &        integer \\
occupant &              Units.People.Age.Group.Decade & categorical &        integer \\
occupant &           Units.People.Age.Group.FiveYears & categorical &        integer \\
occupant &                   Units.People.AirbagUsage & categorical &        integer \\
occupant &             Units.People.AlcoholTestStatus & categorical &        integer \\
occupant &               Units.People.AlcoholTestType & categorical &        integer \\
occupant &              Units.People.AlcoholTestValue &   numerical &        integer \\
occupant &                     Units.People.Condition & categorical &        integer \\
occupant &  Units.People.DOTCompliantMotorcycleHelmet &      binary &        boolean \\
occupant &              Units.People.DriverDistracted & categorical &        integer \\
occupant &                Units.People.DrugTestStatus & categorical &        integer \\
occupant &                  Units.People.DrugTestType & categorical &        integer \\
occupant &                      Units.People.Ejection & categorical &         string \\
occupant &                  Units.People.Endorsements & multi-label &         string \\
occupant &                        Units.People.Gender & categorical &         string \\
occupant &                        Units.People.Injury & categorical &         string \\
occupant &            Units.People.IsAlcoholSuspected &      binary &        boolean \\
occupant &          Units.People.IsMarijuanaSuspected &      binary &        boolean \\
occupant &          Units.People.IsOtherDrugSuspected &      binary &        boolean \\
occupant &            Units.People.OffenseDescription &        text &         string \\
occupant &                       Units.People.OLClass & categorical &        integer \\
occupant &          Units.People.PersonType.Corrected & categorical &         string \\
occupant &           Units.People.SafetyEquipmentUsed & categorical &         string \\
occupant &     Units.People.SeatingPosition.Corrected & categorical &         string \\
occupant & Units.People.SeatingPosition.Corrected.Row & categorical &         string \\
occupant &                       Units.People.Trapped &   numerical &        integer \\
occupant &                    Units.People.UnitNumber &   numerical &        integer \\
 vehicle &                          Units.PostedSpeed &   numerical &        integer \\
 vehicle &                       Units.PreCrashAction & categorical &        integer \\
 vehicle &                    Units.RailGradeCrossing & categorical &        integer \\
 vehicle &                     Units.SequenceOfEvents & multi-label &        integer \\
 vehicle &                      Units.SpecialFunction & categorical &        integer \\
 vehicle &                       Units.TrafficControl & categorical &        integer \\
 vehicle &                       Units.TrafficwayFlow &      binary &        integer \\
 vehicle &                           Units.UnitNumber &   numerical &        integer \\
 vehicle &                             Units.UnitType & categorical &        integer \\
 vehicle &                    Units.UnitType.Verified & categorical &         string \\
 vehicle &                                Units.USDOT &       index &        integer \\
 vehicle &                         Units.VehicleColor & categorical &         string \\
 vehicle &                        Units.VehicleDefect & categorical &        integer \\
 vehicle &                 Units.VehicleMake.Verified & categorical &         string \\
 vehicle &                Units.VehicleModel.Verified & categorical &         string \\
 vehicle &                 Units.VehicleYear.Verified &   numerical &        integer \\
 vehicle &                               Units.Weight & categorical &        integer \\
   crash &                                    Weather & categorical &        integer \\
   crash &                      WithinInterchangeArea &      binary &        boolean \\
   crash &                             WorkersPresent &      binary &        boolean \\
   crash &                            WorkZoneRelated &      binary &        boolean \\
   crash &                               WorkZoneType & categorical &        integer 
\label{tab:full_data_features}
\end{longtable}
}

\section{Post-Processed Features}\label{app:feature}

After the pre-processing steps outlined in Section~\ref{sec:method}, the remaining 62 features are listed below:

\begin{minipage}{\linewidth}
\begin{multicols}{2}
\footnotesize
\begin{itemize}
    \item ActiveSchoolZoneRelated
    \item AnimalRelated
    \item Belted
    \item CrashDate.Month
    \item CrashDate.Time24h
    \item CrashDate.WeekDay
    \item CrashLocationInWorkZone
    \item DividedLaneTravelDirection
    \item DocumentNumber
    \item DriverAge
    \item DriverCondition
    \item DriverDistraction
    \item DriverGender
    \item InCityVillageTownship
    \item IntersectionOrApproachRelated
    \item IsAlcoholRelated
    \item IsCommercialAtFault
    \item IsDrugRelated
    \item LightCondition
    \item LocationPrefix
    \item LocationRoadType
    \item LocationRouteType
    \item Other.Units.UnitType.Verified1
    \item Other.Units.UnitType.Verified2
    \item Other.Units.UnitType.Verified3
    \item Other.Units.UnitType.Verified4
    \item Other.Units.UnitType.Verified5
    \item Other.Units.VehicleModel.Verified1
    \item Other.Units.VehicleModel.Verified2
    \item Other.Units.VehicleModel.Verified3
    \item Other.Units.VehicleModel.Verified4
    \item Other.Units.VehicleModel.Verified5
    \item Other.Units.VehicleYear.Verified1
    \item Other.Units.VehicleYear.Verified2
    \item Other.Units.VehicleYear.Verified3
    \item Other.Units.VehicleYear.Verified4
    \item Other.Units.VehicleYear.Verified5
    \item RoadCondition
    \item RoadContour
    \item RoadSurface
    \item RoadwayDivided
    \item Units.ContributingCircumstance
    \item Units.NumberOfOccupants
    \item Units.NumberOfThruLanes
    \item Units.OccupantsMaxAge
    \item Units.OccupantsMeanAge
    \item Units.OccupantsMinAge
    \item Units.PostedSpeed
    \item Units.PreCrashAction
    \item Units.TrafficControl
    \item Units.TrafficwayFlow
    \item Units.UnitType.Verified
    \item Units.VIN
    \item Units.VehicleColor
    \item Units.VehicleDefect
    \item Units.VehicleMake.Verified
    \item Units.VehicleModel.Verified
    \item Units.VehicleYear.Verified
    \item VIN
    \item VINSeverity
    \item Weather
    \item WorkZoneRelated
\end{itemize}
\end{multicols}
\end{minipage}

\section{JADBio Search Space}\label{app:search_space}
In Table~\ref{tab:search_space} is reported the search space used by the AutoML platform JADBio in this experiment. 
In total, 737 configurations are explored, plus the naive solution.

\begin{table}[tbh]
    \footnotesize
    \centering
    \begin{tabular}{l l l}
        \toprule
        \textbf{Purpose} & \textbf{Algorithm} & \textbf{Hyperparameters} \\
        \midrule
        \multirow{4}{*}{\textit{Feature Selection}} & Epilogi  & threshold = 0.01 \\
        & LASSO & penalty $\in [0, 0.25, 0.5, 0.75, 1.0, 1.25, 1.5, 1.75, 2.0]$ \\
        & Univariate & $\alpha \in  [0.01, 0.001]$ \\
        & SES & $K_{max} \in [2, 3]$, $\alpha \in [0.01, 0.05, 0.1]$ \\
        \midrule
        \multirow{3}{*}{\textit{Predictive Algorithm}} & Ridge LR & $\lambda \in [0.0001, 0.001, 0.1, 1.0, 10, 100]$ \\
        & Decision Tree & $Leaf_{min} \in [1,2,3,4,5]$; $\alpha \in [0.01, 0.05, 0.1]$ \\
        & Random Forest & $N_{trees} \in [100, 1000]$; $Leaf_{min} \in [4,5]$\\
        \bottomrule
    \end{tabular}
    \caption{JADBio search space.}
    \label{tab:search_space}
\end{table}

\section{Complete SHAP analsysis}\label{app:shap_full}

\begin{figure}[tbh]
    \centering
    \includegraphics[width=0.95\linewidth]{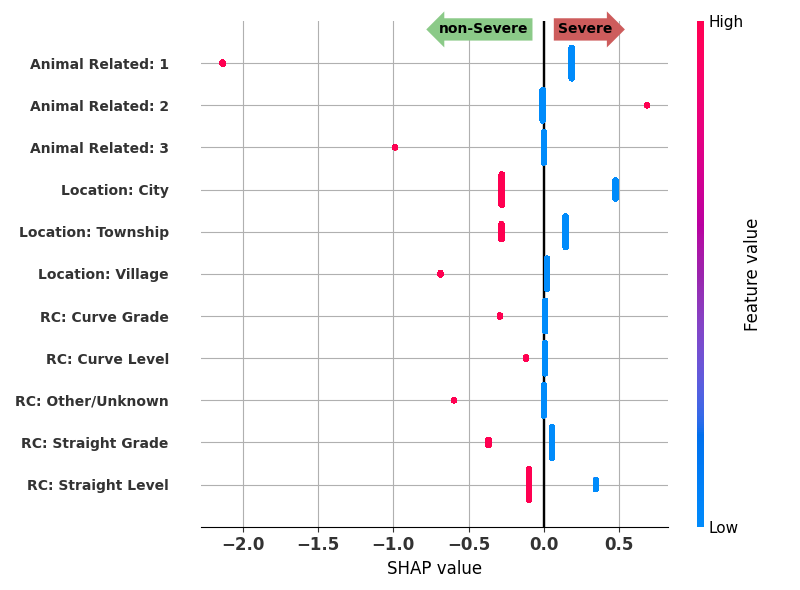}
    \caption{SHAP values corresponding to the environment category with the complete list of features.}
    \label{fig:env_full}
\end{figure}

\begin{figure}[tbh]
    \centering
    \includegraphics[width=0.95\linewidth]{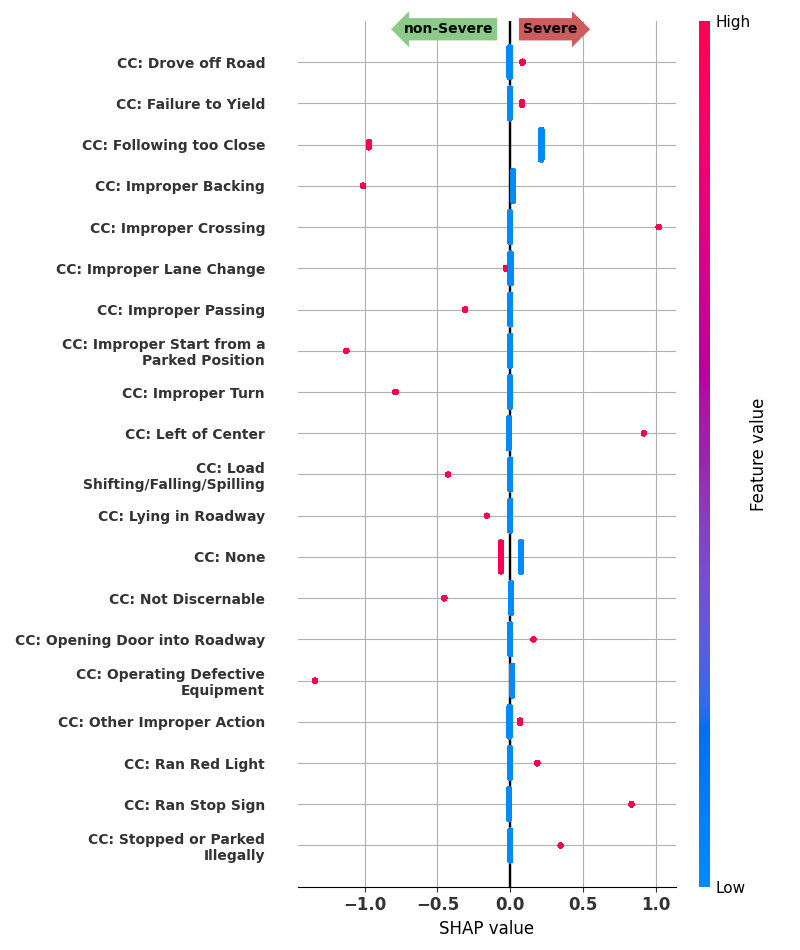}
    \caption{SHAP values corresponding to the operational category with the complete list of features.}
    \label{fig:op_full}
\end{figure}

\begin{figure}[tbh]
    \centering
    \includegraphics[width=0.95\linewidth]{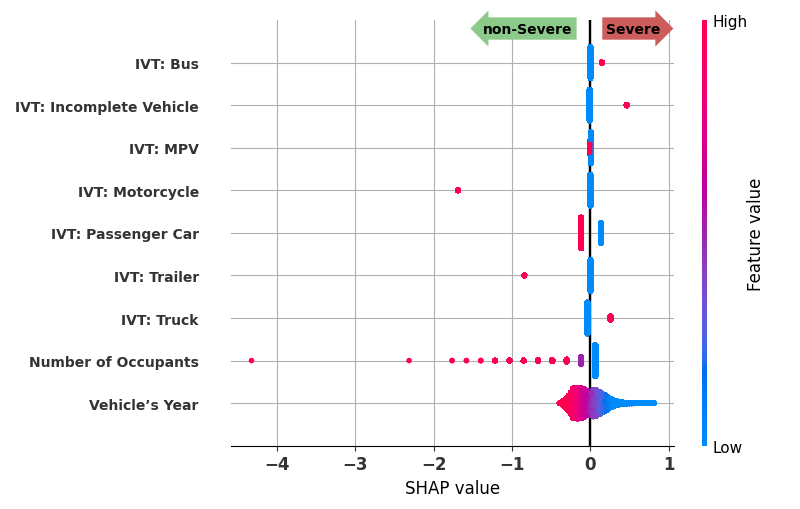}
    \caption{SHAP values corresponding to the vehicle category with the complete list of features.}
    \label{fig:veh_full}
\end{figure}

In Figures~\ref{fig:env_full}, \ref{fig:op_full}, and \ref{fig:veh_full} are reported the SHAP values for the full set of identified features, for the environment, operational, and vehicle categories, respectively.

\clearpage




\section*{Declaration of generative AI and AI-assisted technologies in the writing process}
During the preparation of this work the author(s) used GPT-4.5 in order to improve the readability and language of the work. After using this tool/service, the author(s) reviewed and edited the content as needed and take(s) full responsibility for the content of the publication.

\bibliographystyle{elsarticle-num} 
\bibliography{biblio}

\begin{thebibliography}{10}
\expandafter\ifx\csname url\endcsname\relax
  \def\url#1{\texttt{#1}}\fi
\expandafter\ifx\csname urlprefix\endcsname\relax\def\urlprefix{URL }\fi
\expandafter\ifx\csname href\endcsname\relax
  \def\href#1#2{#2} \def\path#1{#1}\fi

\bibitem{wang2025global}
K.~Wang, Z.~Li, Global, regional, and national road injuries trends from 1990 to 2021: Results from the 2021 {Global Burden of Disease Study}, Injury (2025) 112221.

\bibitem{who}
W.~H. Organization, \href{https://www.who.int/news-room/fact-sheets/detail/road-traffic-injuries}{Road traffic injuries} (2023).
\newline\urlprefix\url{https://www.who.int/news-room/fact-sheets/detail/road-traffic-injuries}

\bibitem{TANDRAYENRAGOOBUR2024}
V.~Tandrayen-Ragoobur, \href{https://www.sciencedirect.com/science/article/pii/S2046043024000339}{The economic burden of road traffic accidents and injuries: A small island perspective}, International Journal of Transportation Science and Technology (2024).
\newblock \href {https://doi.org/https://doi.org/10.1016/j.ijtst.2024.03.002} {\path{doi:https://doi.org/10.1016/j.ijtst.2024.03.002}}.
\newline\urlprefix\url{https://www.sciencedirect.com/science/article/pii/S2046043024000339}

\bibitem{shannon2018applying}
D.~Shannon, F.~Murphy, M.~Mullins, J.~Eggert, Applying crash data to injury claims-an investigation of determinant factors in severe motor vehicle accidents, Accident Analysis \& Prevention 113 (2018) 244--256.

\bibitem{dong2025evaluation}
Y.~Dong, J.~Wood, Evaluation of crash contributing factors (2025).

\bibitem{johansson2009vision}
R.~Johansson, Vision zero--implementing a policy for traffic safety, Safety science 47~(6) (2009) 826--831.

\bibitem{khan2024advancing}
M.~N. Khan, S.~Das, Advancing traffic safety through the safe system approach: A systematic review, Accident Analysis \& Prevention 199 (2024) 107518.

\bibitem{ali2024advances}
Y.~Ali, F.~Hussain, M.~M. Haque, Advances, challenges, and future research needs in machine learning-based crash prediction models: A systematic review, Accident Analysis \& Prevention 194 (2024) 107378.

\bibitem{he2021automl}
X.~He, K.~Zhao, X.~Chu, {AutoM}l: A survey of the state-of-the-art, Knowledge-based systems 212 (2021) 106622.

\bibitem{angarita2021bibliometric}
J.~S. Angarita-Zapata, G.~Maestre-Gongora, J.~F. Calder{\'\i}n, A bibliometric analysis and benchmark of machine learning and automl in crash severity prediction: The case study of three colombian cities, sensors 21~(24) (2021) 8401.

\bibitem{dryadcrash}
A.~Harden, C.~Mary, A.~Castellani, T.~Rodemann, B.~Brian, \href{https://figshare.com/articles/dataset/Comprehensive_multi-level_dataset_of_motor_vehicle_crashes_in_Ohio_USA_2017_2023_Crash_vehicle_and_occupant-level_records_with_detailed_attributes_and_severity_outcomes/29437694}{{Comprehensive multi-level dataset of motor vehicle crashes in Ohio, USA (2017–2023): Crash, vehicle, and occupant-level records with detailed attributes and severity outcomes}} (7 2025).
\newblock \href {https://doi.org/10.6084/m9.figshare.29437694.v1} {\path{doi:10.6084/m9.figshare.29437694.v1}}.
\newline\urlprefix\url{https://figshare.com/articles/dataset/Comprehensive_multi-level_dataset_of_motor_vehicle_crashes_in_Ohio_USA_2017_2023_Crash_vehicle_and_occupant-level_records_with_detailed_attributes_and_severity_outcomes/29437694}

\bibitem{10.1093/bioinformatics/btad545}
K.~Lakiotaki, Z.~Papadovasilakis, V.~Lagani, S.~Fafalios, P.~Charonyktakis, M.~Tsagris, I.~Tsamardinos, Automated machine learning for genome wide association studies, Bioinformatics 39~(9) (2023) btad545.
\newblock \href {http://arxiv.org/abs/https://academic.oup.com/bioinformatics/article-pdf/39/9/btad545/51971215/btad545.pdf} {\path{arXiv:https://academic.oup.com/bioinformatics/article-pdf/39/9/btad545/51971215/btad545.pdf}}, \href {https://doi.org/10.1093/bioinformatics/btad545} {\path{doi:10.1093/bioinformatics/btad545}}.

\bibitem{lundberg2017unified}
S.~M. Lundberg, S.-I. Lee, A unified approach to interpreting model predictions, Advances in neural information processing systems 30 (2017).

\bibitem{li2022extracting}
Z.~Li, Extracting spatial effects from machine learning model using local interpretation method: An example of {SHAP} and {XGBoost}, Computers, Environment and Urban Systems 96 (2022) 101845.

\bibitem{roussou2025investigation}
S.~Roussou, A.~Ziakopoulos, G.~Yannis, Investigation of hit-and-run crash severity through explainable machine learning, Transportation Letters (2025) 1--16.

\bibitem{behboudi2024recent}
N.~Behboudi, S.~Moosavi, R.~Ramnath, Recent advances in traffic accident analysis and prediction: A comprehensive review of machine learning techniques, arXiv preprint arXiv:2406.13968 (2024).

\bibitem{angarita2021case}
J.~S. Angarita-Zapata, G.~Maestre-Gongora, J.~F. Calder{\'\i}n, A case study of {AutoML} for supervised crash severity prediction, in: 19th World Congress of the International Fuzzy Systems Association (IFSA), 12th Conference of the European Society for Fuzzy Logic and Technology (EUSFLAT), and 11th International Summer School on Aggregation Operators (AGOP), Atlantis Press, 2021, pp. 187--194.

\bibitem{baykal2023accident}
T.~Baykal, F.~Ergezer, E.~Eriskin, S.~Terzi, Accident severity prediction in big data using auto-machine learning, Scientia Iranica (2023).

\bibitem{fiorentini2020handling}
N.~Fiorentini, M.~Losa, Handling imbalanced data in road crash severity prediction by machine learning algorithms, Infrastructures 5~(7) (2020) 61.

\bibitem{wen2021applications}
X.~Wen, Y.~Xie, L.~Jiang, Z.~Pu, T.~Ge, Applications of machine learning methods in traffic crash severity modelling: current status and future directions, Transport reviews 41~(6) (2021) 855--879.

\bibitem{sorum2024identification}
N.~G. Sorum, D.~Pal, Identification of the best machine learning model for the prediction of driver injury severity, International journal of injury control and safety promotion 31~(3) (2024) 360--375.

\bibitem{sattar2023transparent}
K.~Sattar, F.~Chikh~Oughali, K.~Assi, N.~Ratrout, A.~Jamal, S.~Masiur~Rahman, Transparent deep machine learning framework for predicting traffic crash severity, Neural Computing and Applications 35~(2) (2023) 1535--1547.

\bibitem{zahid2024factors}
M.~Zahid, M.~F. Habib, M.~Ijaz, I.~Ameer, I.~Ullah, T.~Ahmed, Z.~He, Factors affecting injury severity in motorcycle crashes: Different age groups analysis using {Catboost} and {SHAP} techniques, Traffic injury prevention 25~(3) (2024) 472--481.

\bibitem{shao2024injury}
Y.~Shao, X.~Shi, Y.~Zhang, N.~Shiwakoti, Y.~Xu, Z.~Ye, Injury severity prediction and exploration of behavior-cause relationships in automotive crashes using natural language processing and extreme gradient boosting, Engineering Applications of Artificial Intelligence 133 (2024) 108542.

\bibitem{dong2022predicting}
S.~Dong, A.~Khattak, I.~Ullah, J.~Zhou, A.~Hussain, Predicting and analyzing road traffic injury severity using boosting-based ensemble learning models with {SHAPley} additive {exPlanations}, International journal of environmental research and public health 19~(5) (2022) 2925.

\bibitem{cheng2024crash}
C.~Cheng, S.~Chen, Y.~Ma, F.~Qiao, Z.~Xie, Crash severity prediction and interpretation for road determinants based on a hybrid method, Journal of Transportation Safety \& Security (2024) 1--27.

\bibitem{ostatsDashboard}
{Ohio State Highway Patrol}, \href{https://statepatrol.ohio.gov/dashboards-statistics/ostats-dashboards}{{OSTATS Dashboards}}, \url{https://statepatrol.ohio.gov/dashboards-statistics/ostats-dashboards}, retrieved April 21, 2025 (n.d.).
\newline\urlprefix\url{https://statepatrol.ohio.gov/dashboards-statistics/ostats-dashboards}

\bibitem{NHTSAvindecoder}
{National Highway Traffic Safety Association}, \href{https://vpic.nhtsa.dot.gov/decoder}{{VIN Decoder}}, \url{https://vpic.nhtsa.dot.gov/decoder}, retrieved June 24, 2025 (2025).
\newline\urlprefix\url{https://vpic.nhtsa.dot.gov/decoder}

\bibitem{tsamardinos2022just}
I.~Tsamardinos, P.~Charonyktakis, G.~Papoutsoglou, G.~Borboudakis, K.~Lakiotaki, J.~C. Zenklusen, H.~Juhl, E.~Chatzaki, V.~Lagani, Just {A}dd {D}ata: automated predictive modeling for knowledge discovery and feature selection, NPJ precision oncology 6~(1) (2022) 38.

\bibitem{pmlr-v256-paraschakis24a}
K.~Paraschakis, A.~Castellani, G.~Borboudakis, I.~Tsamardinos, \href{https://proceedings.mlr.press/v256/paraschakis24a.html}{Confidence interval estimation of predictive performance in the context of {AutoML}}, in: K.~Eggensperger, R.~Garnett, J.~Vanschoren, M.~Lindauer, J.~R. Gardner (Eds.), Proceedings of the Third International Conference on Automated Machine Learning, Vol. 256 of Proceedings of Machine Learning Research, PMLR, 2024, pp. 4/1--14.
\newline\urlprefix\url{https://proceedings.mlr.press/v256/paraschakis24a.html}

\bibitem{lagani2016feature}
V.~Lagani, G.~Athineou, A.~Farcomeni, M.~Tsagris, I.~Tsamardinos, Feature selection with the r package mxm: Discovering statistically-equivalent feature subsets, arXiv preprint arXiv:1611.03227 (2016).

\bibitem{christoph2020interpretable}
M.~Christoph, Interpretable machine learning: A guide for making black box models explainable (2020).

\bibitem{safari2020comprehensive}
M.~Safari, S.~S. Alizadeh, H.~S. Bazargani, A.~Aliashrafi, A.~Maleki, P.~Moshashaei, M.~Shakerkhatibi, A comprehensive review on risk factors affecting the crash severity, Iranian journal of health, safety and environment 6~(4) (2020) 1366--1376.

\bibitem{newgard2008defining}
C.~D. Newgard, Defining the “older” crash victim: The relationship between age and serious injury in motor vehicle crashes, Accident Analysis \& Prevention 40~(4) (2008) 1498--1505.

\bibitem{ryan1998age}
G.~A. Ryan, M.~Legge, D.~Rosman, Age related changes in drivers' crash risk and crash type, Accident Analysis \& Prevention 30~(3) (1998) 379--387.

\bibitem{de2010driver}
J.~C. de~Winter, D.~Dodou, The driver behaviour questionnaire as a predictor of accidents: A meta-analysis, Journal of safety research 41~(6) (2010) 463--470.

\bibitem{febres2020influence}
J.~D. Febres, S.~Garc{\'\i}a-Herrero, S.~Herrera, J.~Guti{\'e}rrez, M.~A. Mariscal, Influence of seat-belt use on the severity of injury in traffic accidents, European transport research review 12~(1) (2020) 9.

\bibitem{furlan2020advanced}
A.~D. Furlan, T.~Kajaks, M.~Tiong, M.~Lavalli{\`e}re, J.~L. Campos, J.~Babineau, S.~Haghzare, T.~Ma, B.~Vrkljan, Advanced vehicle technologies and road safety: A scoping review of the evidence, Accident Analysis \& Prevention 147 (2020) 105741.

\end{thebibliography}

\end{document}